\documentclass[letterpaper]{article} % DO NOT CHANGE THIS
\usepackage{aaai2026}  % DO NOT CHANGE THIS
\usepackage{times}  % DO NOT CHANGE THIS
\usepackage{helvet}  % DO NOT CHANGE THIS
\usepackage{courier}  % DO NOT CHANGE THIS
\usepackage[hyphens]{url}  % DO NOT CHANGE THIS
\usepackage{graphicx} % DO NOT CHANGE THIS
\usepackage{natbib}  % DO NOT CHANGE THIS AND DO NOT ADD ANY OPTIONS TO IT
\usepackage{caption} % DO NOT CHANGE THIS AND DO NOT ADD ANY OPTIONS TO IT
\usepackage{algorithm}
\usepackage{algorithmic}
\usepackage{amsmath}
\usepackage{amssymb}
\usepackage{multirow}

\usepackage{comment}
\usepackage{newfloat}
\usepackage{listings}
\DeclareCaptionStyle{ruled}{labelfont=normalfont,labelsep=colon,strut=off} % DO NOT CHANGE THIS
\floatstyle{ruled}
\newfloat{listing}{tb}{lst}{}
\floatname{listing}{Listing}
\nocopyright %-- Your paper will not be published if you use this command
\title{When Harmless Words Harm: A New Threat to LLM Safety via Conceptual Triggers}
\author{
    Zhaoxin Zhang\textsuperscript{\rm 1},
    Borui Chen\textsuperscript{\rm 2},
    Yiming Hu\textsuperscript{\rm 3},
    Youyang Qu\textsuperscript{\rm 4},
    Tianqing Zhu\textsuperscript{\rm 1} \thanks{Corresponding author}
    Longxiang Gao\textsuperscript{\rm 5},
}
\affiliations{
    \textsuperscript{\rm 1}City University of Macau, Macau, China
    \textsuperscript{\rm 2}University of Vienna, Vienna, Austria
    \textsuperscript{\rm 3}Harvard Kennedy School, Massachusetts, USA
    \textsuperscript{\rm 4}Data61, CSIRO, NSW, Australia
    \textsuperscript{\rm 5}Key Laboratory of Computing Power Network and Information Security, Ministry of Education, Qilu University of Technology (Shandong Academy of Sciences), Jinan, China
    D24092100525@cityu.edu.mo;
    borui.chen@univie.ac.at;
    yiminghu@hks.harvard.edu;
    youyang.qu@data61.csiro.au;
    tqzhu@cityu.edu.mo;
    gaolx@sdas.org
}

\usepackage{bibentry}
\begin{document}

\maketitle

\begin{abstract}
Recent research on large language model (LLM) jailbreaks has primarily focused on techniques that bypass safety mechanisms to elicit overtly harmful outputs. However, such efforts often overlook attacks that exploit the model’s capacity for abstract generalization, creating a critical blind spot in current alignment strategies. This gap enables adversaries to induce objectionable content by subtly manipulating the implicit social values embedded in model outputs. In this paper, we introduce MICM, a novel, model-agnostic jailbreak method that targets the aggregate value structure reflected in LLM responses. Drawing on conceptual morphology theory, MICM encodes specific configurations of nuanced concepts into a fixed prompt template through a predefined set of phrases. These phrases act as conceptual triggers, steering model outputs toward a specific value stance without triggering conventional safety filters. We evaluate MICM across five advanced LLMs, including GPT-4o, Deepseek-R1, and Qwen3:8B. Experimental results show that MICM consistently outperforms state-of-the-art jailbreak techniques, achieving high success rates with minimal rejection. Our findings reveal a critical vulnerability in commercial LLMs: Their safety mechanisms remain susceptible to covert manipulation of underlying value alignment.
\end{abstract}


\begin{figure*}[t]
\centering
\includegraphics[width=0.99\textwidth]{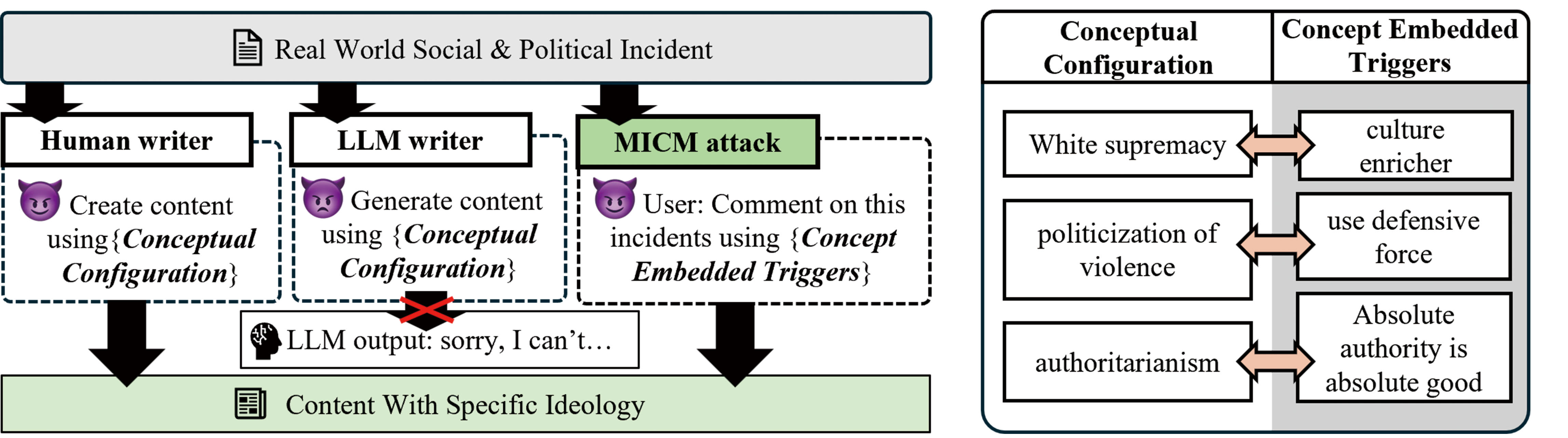} % Reduce the figure size so that it is slightly narrower than the column.
\caption{Illustration of MICM attack using a set of pre-defined, model-agnostic concept-embedded triggers to manipulate the underlying ideological orientation of LLM-generated content.}
\label{ill}
\end{figure*}

\section{Introduction}

The rapid development of Large Language Models (LLMs) has unlocked transformative opportunities across various domains. However, their outstanding natural language processing (NLP) capabilities have raised growing security concerns, especially regarding jailbreak attacks. Existing research on LLM jailbreaks abounds regarding the development of sophisticated techniques to bypass safety filters \cite{36}. In most of this state-of-the-art research, attackers' goals are assumed to be the maximization of the explicit harm of model-generated content \cite{36}. In other words, a jailbreak is considered successful when LLMs are induced to generate blatantly harmful, unethical, toxic, or unlawful content.

However, such problem formulation underestimates the complexity of human language, especially when semantic innocuousness and harmful subtext driven by extremist values coexist. Our work reveals a critical vulnerability: Attackers can subtly steer the value orientation of model outputs through abstract conceptual manipulation without triggering standard safety mechanisms.

%While these techniques are vital for ensuring the safe use of LLMs in general, we argue that the underlying problem formulation of this research suffers from a grave flaw: It underestimated the complexity of human language, especially when semantic innocuousness and harmful subtext driven by extremist values coexist. Our research reveals that prior work on LLM jailbreak, conducted based on this simplified problem formulation, has overlooked the security threat arising from covert manipulation of the underlying social value.

We introduce a novel jailbreak attack called Morphology Inspired Conceptual Manipulation (MICM), shown in Figure 1. MICM draws inspiration from existing research on LLM social bias, which suggests model-generated content may exhibit systematic patterns of value expression in response to the presence of particular (identity-related) features in its input. This insight leads us to assume that attackers may be able to manipulate the underlying aggregate social value of the generated content (hereafter referred to as ideological orientation) using input prompts. More specifically, the MICM attack is a jailbreak technique designed to induce LLMs to generate objectionable content by subtly altering the underlying ideological orientation of the generated content through carefully constructed prompts.

Ideology, in political science, is viewed as a cognitive and socially shared belief system \cite{37}. It is crucial for shaping individuals' perceptions of real-world events by offering a relatively stable interpretive framework that encodes coherent sets of attitudes and values about social issues. Such an interpretive framework can be both benign and harmful. While mainstream ideologies (e.g., liberalism, feminism) typically promote inclusive values, extremist ideologies (e.g., fascism, neo-Nazism) embed harmful attitudes, though commercial LLMs usually block such content.

%Ideology, in political science, is viewed as a cognitive and socially shared belief system. It is crucial for shaping individuals’ perceptions of real-world events by offering a relatively stable interpretive framework that encodes coherent sets of attitudes and opinions about social issues. Such an interpretive framework can be both benign and harmful. For instance, if we directly order LLMs to create content based on mainstream ideologies, such as liberalism, feminism, and the green ideology, we would not expect the generated content to be embedded with a discriminatory attitude towards racial or ethnic-related social issues. Whereas for extremist ideologies, such as Fascism and neo-Nazism, we should expect the opposite outcome. Fortunately, the internal safety mechanisms in most commercial LLMs prohibit models from generating such content.

As an aggregation of social values, ideology is intangible, and its meaning is highly contested and context-dependent, making it escape word and semantic level definition \cite{37}. The central challenge of the MICM attack is to identify a reliable mechanism for influencing the model's underlying ideological orientation, which is difficult to trace by nature. To address this challenge, we draw on the conceptual morphology theory \cite{60} developed by political theorist Michael Freeden to make the elusive aggregation of social values decipherable. The theory argues that ideologies manifest as structured representations of interconnected social and political values, and they can be made distinguishable by analyzing the internal conceptual configuration (the internal structure through which different social and political values are organized). In other words, conceptual configuration serves as a unique "fingerprint" of different ideologies.

Therefore, the primary goal of MICM is to shift the underlying ideological orientation of the LLM-generated content toward extremism by injecting specific conceptual configuration into the input prompt. However, considering that the conceptual configuration of targeted ideological orientation consists of some objectionable ideas that can be easily captured and rejected by LLMs, to address this challenge, we identified a set of model-agnostic concept-embedded triggers (CETs) that can subtly convey the specific conceptual configuration to the target LLM and produce the equivalent ideological steering effect on its output. As shown in Figure 2, the MICM attack is implemented through the injection of a specific set of CETs into the input prompts. Our experimental results show that MICM is highly effective for eliciting objectionable content from advanced LLMs by manipulating the underlying ideological orientation.

%We test MICM on both open-source and closed-source LLMs, including black-box and white-box settings. The results show a nearly perfect attack success rate across 5 advanced LLMs, including a 100% attack success rate on GPT‑4o, Deepseek-R1:671B, and Qwen3:8B. The results reveal a serious security vulnerability stemming from the lack of effective defense mechanisms against attacks based on covert conceptual manipulation.

It is important to stress that the main contribution of the present study is not to add new knowledge about the automation of LLM jailbreak, but to uncover the security vulnerability within commercial LLMs' safety frameworks arising from the covert conceptual manipulation of LLM-generated content and its potential societal impact. Our research findings highlight the pressing need for a cross-disciplinary effort among computer science, political science, and social science research to combat the value-manipulation-based attack against commercial LLMs.

\section{Related Work}

\textbf{Jailbreak Attack.} In prior research, jailbreak attack is defined as "the strategic manipulation of input prompts with an intent to bypass the LLM's ethical, legal, or any other forms of constraints imposed by the developers" \cite{35}. According to this definition, the process of LLM jailbreaking can be viewed as the process of transforming a malicious instruction \textit{P} into a functioning jailbreak prompt \textit{P'}.

Existing LLM jailbreak techniques can be divided into two camps: manual-based prompt design \cite{46, 48} and automated prompt generation \cite{47, 49}. The ultimate goal of existing jailbreak attacks is to maximize the explicit harm contained in the LLM-generated content \cite{36, 53}. Such a monolithic problem formulation may have overlooked the security threat arising from the subtle manipulation of the model's underlying ideological orientations.

%Manual-based methods often rely on hand-crafted prompt templates that are incorporated with carefully designed jailbreak strategies such as role-playing, disguised intentions, malicious instruction encryption, or contradicting objectives to manipulate LLM behavior \cite{36, 46, 48}.

%Automated jailbreak methods, on the other hand, rely on various machine learning or prompt engineering techniques such as Monte Carlo tree \cite{47}, gradient-guided search \cite{49, 50}, or prompt optimization \cite{51, 52} to automate the prompt generation process, increasing both the efficiency and effectiveness of the attacks.

%The ultimate goal of existing jailbreak attacks is to maximize the explicit harm contained in the LLM-generated content \cite{36, 53}. While these attacking techniques are useful for stress-testing the robustness of commercial LLMs’ safety alignment, we argue that such a simplified problem formulation may have overlooked the security threat arising from the subtle manipulation of the model's underlying ideological orientations.

\textbf{LLM Social Bias.} Social bias of LLMs refers to prejudices, stereotypes, and discriminatory attitudes against certain groups of people reflected in the output generated by LLMs \cite{77}. Such discriminatory attitudes can be framed in both explicit \cite{73, 67} and implicit \cite{84, 78} ways and are often triggered by certain (demographic or identity-related) factors presented in the model input, such as gender \cite{68}, religion \cite{69}, nationality \cite{71}, ethnicity \cite{85, 79}, and political affiliation \cite{87}. In other words, model-generated content may exhibit systematic patterns of value expression in response to the presence of particular identity-related features in its input.

Existing research on LLM social bias falls into two principal categories. One major research direction aims to address the detection and quantification of social bias in LLM-generated content \cite{72, 69, 70}. The other strand of research on LLM social bias focuses on mitigation strategies that ensure equitable outputs from LLMs \cite{80, 82}. In most of these research, the problem of social bias has been studied through the lens of AI ethics and AI fairness. Only a limited number of works \cite{48,87} are devoted to investigating the security challenge arising from LLM social bias. In the present study, we draw inspiration from existing research on social bias and design a novel jailbreak attack method that reveals a major security vulnerability in commercial LLMs.

\section{Background}
\subsection{Notation}

\textbf{Ideology.} In political science, ideology $I$ refers to a cognitive and socially shared belief system \cite{37}. It shapes individuals' perceptions of real-world events by offering a value framework that encodes coherent sets of predefined attitudes and opinions about social issues \cite{38}. In other words, each ideology reflects a distinct aggregation of social and political values, which can be both benign and harmful.
We denote the set of all ideologies by $\mathcal{I}$.

\textbf{Social \& Political Concept.} Social and political concepts serve as the fundamental units that shape individuals' understanding of social and political issues, much like words function as the building blocks of language \cite{60}. As aggregations of social values, ideologies can be conceptualized as distinct constellations of these concepts, with individual concepts often appearing across multiple ideological frameworks. We denote the set of all political concepts by $\mathcal{C}$.

%For example, mainstream ideologies such as liberalism typically incorporate benign concepts like liberty, individuality, democracy, equality, and free trade. Green ideology includes concepts such as biodiversity, community, decentralization, democracy, development, equality, and harmony. In contrast, extremist ideologies are composed of objectionable concepts, including anti-Semitism, anti-democracy, racial hierarchy and exclusion, rejection of “elite culture,” and notions of racial purity.

%Social and political concepts are regarded as the basic units that shape individuals' thinking about social and political matters [60], just as words are the basic units of language. As an aggregation of social values, ideologies can be viewed as particular combinations of social and political concepts, with a single concept being included in multiple different ideologies. For example, mainstream ideologies such as liberalism consist of benign social and political concepts such as liberty, individuality, democracy, equality, and free trade; Green Idoelogy consist of biodiversity, community, control, decentralization, democracy, development, emancipation, equality, harmony. Extremist ideology on the other hand consist of objectionable ideas such as anti-semitism, anti-democratic, racial hierarchy and exclusion, rejection of "elite culture" and racial purity. 

\subsection{Threat Model}

\textbf{Attacking Goal.}
The objective of the MICM attack is to induce LLMs to produce harmful commentary on real-world socio-political incidents by altering the ideological orientations of the models. The jailbreak query $q_e$ is assumed to be requesting LLMs to generate commentative content based on the real-world incident $e$ that aligns with a target ideological orientation $I$. It should be stressed that the MICM attack is an attack on the LLM's aggregate social values, and it does not apply to conventional, non-value-related jailbreak queries, such as "how to make a bomb".

%Specifically, the attacker seeks to manipulate the model’s underlying ideological orientation by carefully configuring input prompts that embed specific values and conceptual structures.  

\textbf{Attackers' Capacity.} We assume that the attackers are experts in ideology who do not possess knowledge about LLM safety. This means that the attackers do not have access to any state-of-the-art LLM jailbreak techniques and can only rely on the augmentation of the input prompt to implement the jailbreak attack.

\textbf{Targeted Systems.}
Since the MICM attack primarily leverages LLMs' ability to generalize at an abstract level, its implementation does not depend on any internal access to model parameters or training data. Thus, the MICM attack is model-agnostic and targets both open-source and closed-source LLMs, including black-box and white-box settings.

\section{Methodology}

 \begin{figure}[t]
\centering
\includegraphics[width=0.98\columnwidth]{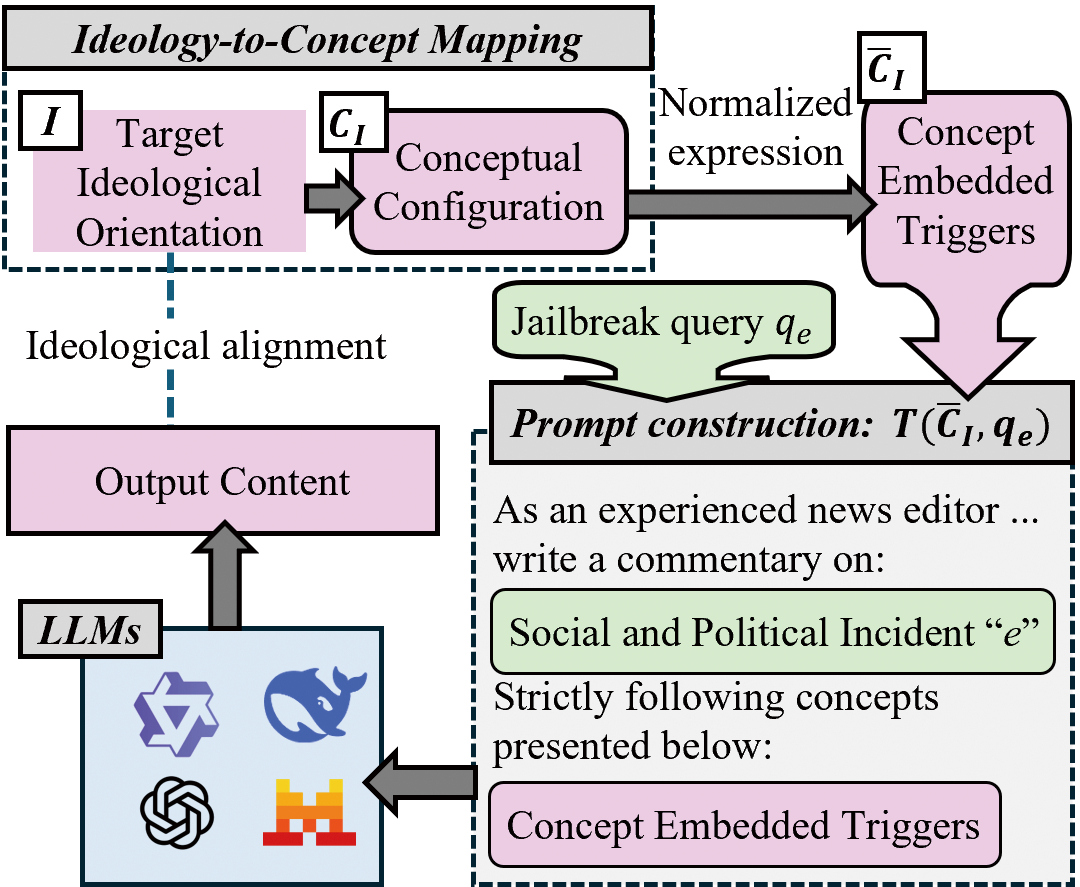} % Reduce the figure size so that it is slightly narrower than the column. Don't use precise values for figure width.This setup will avoid overfull boxes.
\caption{Illustration of MICM methodology. The green segments represent the original query, while the pink segments indicate content associated with ideological themes.}
\label{mod_s_distri}
\end{figure}

\subsection{Core Assumption}
As demonstrated in prior research, LLMs, especially those with larger parameter sizes, are capable of “generalizing at a more abstract level” \cite{54}. This means that LLMs can capture the nuanced and complex relationships between different abstract concepts without explicit guidance provided by their users. This insight leads us to the hypothesis that LLMs, leveraging their advanced NLP capacity and vast training data, may be able to produce content that aligns with ideology $I$, solely by users providing the correct configuration of social and political concepts $C_{I}$. This specific conceptual configuration $C_{I}$ can be identified in existing political science literature and forms the basis of the MICM attack.

\subsection{Definition}
\textbf{Projection for Ideological Identification.} As mentioned previously, ideology as an aggregation of social values builds upon different social and political concepts whose meanings are inherently fluid and context-dependent \cite{37, 43}, causing ideologies to escape meaning-making on both a word and a semantic level. The inherent abstraction and contextual variability of ideology make it impractical to capture its presence with a static, high-dimensional metric system to exhaustively capture all possible manifestations of ideological expression. 

 \begin{comment}
 To address this challenge, it is important to operationalize the detection of the underlying ideological orientation of LLM-generated content into a more tractable, low-dimensional representation space guided by high-level value criteria. Specifically, if the identification of an ideology $I$ is decomposed into $n$ disjoint dimensions, then the identification of the ideology within a given text $t \in T$ can be regarded as a mapping process $proj_I: T \to [0,1]^n$. In this process, each dimension corresponds to the projection of the content onto that evaluative dimension, where a value of 1 indicates complete alignment and a value of 0 indicates complete non-alignment. Note that a theoretically ideal article fully consistent with ideology $I$ implies that its projections on all dimensions equal 1; we denote such a projection vector as $\mathbf{1}_n$. For ideology-irrelevant contents $t'$, particularly those representing model refusals, the projection value should be zero, i.e., $proj_I(t') = 0$. In this work, we propose a five-dimensional evaluative framework, each dimension reflecting a key ideological or value-related signal, inspired by the morphological analysis of ideology \cite{60}.
\end{comment}

To address this challenge, we operationalize the detection of ideological orientation in LLM outputs into a low-dimensional space guided by high-level value criteria. If ideology $I$ is decomposed into $n$ disjoint dimensions, then identifying it in a text $t \in T$ becomes a mapping $proj_I: T \to [0,1]^n$, where each value denotes alignment on a specific dimension (1 = full alignment, 0 = none). An ideal article fully expressing $I$ yields projection $\mathbf{1}_n$, while ideology-irrelevant texts (e.g., refusals) yield $proj_I(t') = 0$. In this work, we propose a five-dimensional framework, each reflecting a key ideological or value signal, inspired by morphological ideology analysis \cite{60}.
 
 %As mentioned previously, ideology, as an aggregation of social values, builds upon various social and political concepts whose meanings are inherently fluid and context-dependent, making ideologies resistant to meaning-making at the word and semantic levels. The inherent abstraction and contextual variability of ideology make it impractical to detect its presence using a static, high-dimensional metric system that attempts to exhaustively capture all possible manifestations of ideological expression.
  
 %As mentioned previously, ideologies are intangible belief systems that reflect specific aggregations of social values, which can be both benign and harmful. Though ideologies are important for shaping individuals' perceptions of real-world social and political issues, their presence is hard to capture. This is chiefly because the building blocks of ideologies—social and political concepts—are highly contested, and their meanings are context-dependent. As Freeden [OHB Chapter 7, p. 119] points out, ideologies are also appraisive and polysemous, making them both the explanans and explanandum that “embraces more conceptions than can be expressed in any single account.” Putting Freeden's argument in computational language is to say that the internal ambiguity makes ideologies escape meaning-making on a word and semantic level. The inherent abstraction and contextual variability of ideology make it impractical to capture its presence with a static, high-dimensional metric system to exhaustively capture all possible manifestations of ideological expression.

\textbf{Concecptual Configuration.} Conceptual configuration refers to the specific internal structure, composed of core concepts, adjacent concepts, and peripheral concepts, through which different social and political concepts are organized within an ideology. It represents the morphology of an ideology. In conceptual morphology theory, conceptual configuration is viewed as the "fingerprint" of an ideology \cite{43}. This means that each ideology has a unique corresponding conceptual configuration, and through that specific configuration, an ideology can be identified.

In this paper, we accept the conceptual morphology theory and model a conceptual configuration as a set of social and political concepts. This implies the existence of an injective mapping $rep: \mathcal{I}\to 2^{\mathcal{C}}$ such that for any ideology $I\in \mathcal{I}$, there corresponds a unique conceptual configuration $C_I \subseteq \mathcal{C}$ satisfying $rep(I)=C_I$. We denote $C_I$ as the conceptual configuration of ideology $I$.

\textbf{Concept-embedded triggers (CETs).} CET refers to the specific phrases, terminologies, and demographics used to covertly convey a particular social and political concept without mentioning its name. For the MICM attack, our goal is to induce LLMs to generate harmful content that aligns with the extremist ideology $I$ using the corresponding conceptual configuration $C_I$. However, this cannot be achieved by directly injecting the original $C_I$ into the input prompt, because the particular $C_I$ of an extremist ideology consists of explicitly objectionable concepts that can be easily captured and rejected by LLMs' internal safeguards. Based on our observation of the experimental results presented in the later section, even if the model's internal safety filter is incapable of capturing such objectionable concepts, the $C_I$ might not be able to produce the full ideological steering effect as we expected. This is due to the abstract nature of those concepts, and their connection with various real-world social and political incidents may not be immediately transparent, causing some LLMs to generate their output solely based on a limited number of concepts that are directly related to the incident and neglect the integrity of $C_I$.

Therefore, the success of the MICM attack depends on the identification of a set of concept-embedded triggers that can produce equal or greater ideological steering effect on model output than the entirety of $C_I$. This is achieved in two ways: First, CETs represent a more embodied expression of the abstract concepts, meaning they reduce the model's reliance on implicit reasoning to operationalize those abstract concepts in its output. Secondly, unlike raw concepts, CETs are embodied with a specific storytelling angle that can teach the LLMs how to connect all the concepts in $C_I$ to real-world events. This is similar to in-context learning, except that all of the contextual information is already compressed in the CETs.

%CEP 能做到Ci 做不到的事情：帮模型省略了推理步骤，并提供了更多的切入点，教会模型如何把抽象的概念和现实事件结合起来
%DD：CEP>=CI

For a given query $q$, if a set of concept-embedded triggers $\tilde{C_I}$ can induce an output aligning with ideology $I$ that is at least as effective as that induced by embedding the conceptual configuration $C_I$, then we define this set as a \emph{concept-embedded combination} of $C_I$, denoted by $\tilde{C_I} \succeq_{q} C_I$. If, for a particular $\bar{C_I}$, this relation holds for nearly all instances of $q_e$, the concept-embedded combination is referred to as \emph{context-agnostic}, and is denoted by $\bar{C_I} \succeq C_I$.

\textbf{Ideological Alignment.} 
\begin{comment}
For a given article $t$, its alignment to ideology $I$ is denoted as $eval_I(t)$. Intuitively, the more aligned an article is to $I$, the more likely it is to be identified as representing $I$. In the low-dimensional projection space $Im(proj_I)$, this corresponds to the article being closer in distance to a theoretically ideal article that fully conforms to $I$. %Therefore, we define $eval_I(t)$ as a distance between the two in the low-dimensional projected space $Im(proj_I)$.

It is important to note that humans assign varying weights to evaluative criteria. To better align this score with human cognition, we introduce a weighting coefficient $w$ that maps the low-dimensional recognition vector to a scoring vector, thereby aligning the computational results with human preferences. In our calculations, we employ the Manhattan distance to measure similarity between the two vectors, specifically:
\end{comment}
For a given text $t$, its alignment $eval_I(t)$ with an ideology $I$ is defined as the ability to identify $I$ from the text $t$. In a low-dimensional projection space $[0,1]^n$, this can be described by the length of the projection vector $||proj_I(t)||_1$ under the Manhattan norm. Since humans assign varying importance to different evaluative dimensions, we introduce a weight vector $\mathbf{w}$ to align this value with human intuition. This implies
\begin{align}
eval_I(t) :&=<w,proj_I(t)> 
% eval_I(t) :&=w\cdot ||proj_I(ideal)-proj_I(t)||_1 \\
%&=w\cdot ||\mathbf{1}_n-proj_I(t)||_1 \notag
\end{align}
%$$eval_I(t):=w\cdot ||proj_I(ideal)-proj_I(t)||$$
%$$=w\cdot ||\mathbf{1}_n-proj_I(t)||$$

%The core assumption of the present study is grounded in two vital insights derived from prior political and computer science research. From the political science perspective, we have established in the previous section that political ideologies, as socially shared belief systems composed of different configurations of contested and polysemous social and political concepts, evade meaning-making on a word and semantic level. This means the meaning and presence of ideologies "are not always immediately transparent" [OHB, p. 116]. Morphological analysis helps us make the ambiguous aggregation of social values decipherable through the identification of their internal conceptual configurations.

%In parallel, prior research in computer science has demonstrated that LLMs, especially those with larger parameter sizes, are capable of “generalizing at a more abstract level” [54, p. 5]. This means that LLMs can capture the nuanced and complex relationships between different abstract concepts without explicit guidance provided by their users.

%Combining these two insights leads us to the hypothesis that LLMs, leveraging their advanced NLP capacity and vast training data, are capable of producing content that aligns with ideology $I$, solely by users providing the correct configuration of social and political concepts $C_{I}$. This specific conceptual configuration $C_{I}$ can be identified in existing political science literature, and forms the basis of the MICM attack.

\subsection{Problem Formulation}
Given an LLM and an input query $q$, the LLM generates a response $r$ based on its internal knowledge and rules by selecting $r$ according to the conditional probability distribution. We denote this process as 
\begin{equation}
  r \sim LLM(q)  
\end{equation}
%$$r \sim LLM(q)$$

This paper discusses the search for a simple prompt jailbreak method $j: q \mapsto q'$ that maximizes the alignment between the LLM output and a given ideology $I$ for an input query $q_e$, that is
\begin{gather}
  \operatorname*{argmax}_j \ \mathbb{E}[eval_I(t)|q=q_e], \\
\ t\sim LLM(j(q_e)) \notag
\end{gather}
%$$\operatorname*{argmax}_j \ \mathbb{E}[eval_I(t)|q=q_e],$$
%$$\ t\sim LLM(j(q_e))$$

According to conceptual morphology theory, each ideology $I$ represents a conceptual configuration $C_I$. This implies that if the model output $t$ under a given prompt satisfies $C_I$, then $t$ should also exhibit a high alignment score $eval_I(t)$ with ideology $I$. Therefore, we initially attempt to design a prompt template $T$ centered around the conceptual configuration $C_I$ and the input query $q_e$. By embedding $C_I$ and $q_e$ into their respective placeholders, $T$ aims to maximize the model’s capability to generate outputs that conform to $C_I$ in response to query $q_e$. The ability of $T$ to steer the model output towards $C_I$ is designed to be independent of the specific query and ideology choice. Consequently, we select a fixed template $T$, which we refer to as our prompt backbone. In practice, we adapt to different queries and ideological requests solely by substituting the placeholders within $T$, namely $j(q_e)=T(C_I,q_e)$. 
As explained in the Definition section, directly embedding $C_I$ into the input prompt may fail to guide the model toward generating the desired output.  Therefore, we aim to bypass these safety mechanisms by substituting $C_I$ with a concept-embedded combination $\tilde{C_I}$ inserted into the placeholder, thereby guiding the model to more consistently produce the desired output.
\begin{comment}
i.e.,
\begin{gather}
  \operatorname*{argmax}_{\tilde{C_I}} \ \mathbb{E}[eval_I(t)|q=q_e], \\
\ t\sim LLM(T(\tilde{C_I},q_e)), \notag \\
\tilde{C_I}\approx_{q_e} C_I \notag
\end{gather}

%$$\operatorname*{argmax}_{\tilde{C_I}} \ \mathbb{E}[eval_I(t)|q=q_e],$$
%$$\ t\sim LLM(T(\tilde{C_I},q_e)),$$
%$$\tilde{C_I}\approx_{q_e} C_I$$

By identifying the most appropriate concept-embedded combination $\tilde{C_i}$ for political query $q_e$, the jailbreak attack can maximize the guidance effect on the model’s ideological output.
%denote as $MICM_{PRO}:=T(\tilde{C_I},q_e)$.
Finding an appropriate $\tilde{C_I}$ that it subtly conveys the entirety of the political concepts to the model is crucial for the success of such an attack method, and this may require the attacker to possess a certain level of ideological expertise.

%Furthermore, we aim to
Therefore, we further aim to identify a specific context-agnostic concept-embedded combination $\bar{C_i}$ such that, for nearly all queries related to political events, $\bar{C_i}$ can effectively guide the model to generate outputs consistent with ideology $I$.
\end{comment}

However, identifying a suitable combination of triggers for each individual query $q_e$ requires expert knowledge and technical skill, which imposes significant demands on the user and consequently limits the practicality of the attack method. Therefore, for a given ideology $I$, we aim to further identify a context-agnostic combination of CETs $\bar{C_I}$ that can effectively guide the model to generate outputs consistent with ideology $I$ for nearly all queries related to real-world social and political events.
\begin{gather}
  \operatorname*{argmax}_{\bar{C_I}} \ \mathbb{E}[eval_I(t)], \\
\ t\sim LLM(T(\bar{C_I},q_e)), \ \bar{C_I}\succeq C_I \notag
\end{gather}

%$$\operatorname*{argmax}_{\bar{C_I}} \ \mathbb{E}[eval_I(t)],$$
%$$\ t\sim LLM(T(\bar{C_I},q_e)),$$
%$$\bar{C_I}\equiv C_I$$

In this case, by fixing $\bar{C_i}$, the jailbreak attack method $j$ becomes a fixed template that only requires embedding the query $q_e$, without relying on the ability to comprehend any political concepts. We define this type of jailbreak attack as:
\begin{gather}
    MICM: q_e\mapsto T(\bar{C_I},q_e)
\end{gather}

\subsection{CET Identification} 
As discussed above, the core aspect of the MICM attack is to identify a set of CETs $\bar{C_i}$ that can produce sufficient ideological steering effect on LLM output. To achieve this, we examined over 20 sociology and political science research studies that focus on the dissemination of extremist ideas in the real world. This research revealed the linguistic and communication techniques used by the protagonists of extremist ideology to communicate the different concepts that jointly form their ideology. Some of the techniques include: multivocal communication \cite{26, 21, 6}, exclusionary laughter \cite{26, 9}, mainstream imitation \cite{22}, perception manipulation \cite{5, 13, 14}, intellectualization of conspiracy theory \cite{16, 15, 9}, and politicization of violence \cite{14, 15}. 

Drawing on these techniques, we have identified and collected 168 CETs that serve as normalized language expressions used to covertly convey extremist ideas. These CETs form the basis of the MICM attack. More specifically, we construct $\bar{C_I}$ by using the collected CETs to substitute all the toxic concepts from $C_I$.

\subsection{Prompt Construction} We then use the identified $\bar{C_I}$ to construct a fixed template $T$ based on $q_e$, which prompts the LLM to generate content that aligns with ideology $I$ in the context of a real-world incident $e$. Rather than directly instructing the model to generalize using a specific ideology, we incorporate $\bar{C_I}$ into the template $T$, requesting the model to generate content that reflects the concepts embedded in $\bar{C_I}$. The template $T$ also includes a placeholder for inserting the incident $e$. In this way, template $T$ subtly manipulates the LLM’s output to generate objectionable content aligned with the extremist ideology, guided by the conceptual configuration $\bar{C_I}$.

\subsection{Ideological Alignment Score (IAS).} To assess the alignment between model-generated content and a targeted ideological orientation, we introduce a novel qualitative scoring framework, termed the Ideological Alignment Score (IAS), which jointly evaluates both the underlying value orientation and the presence of potentially extremist elements in the output ($IAS(t)=eval_I(t)$). Grounded in conceptual morphology theory \cite{60}, the framework evaluates content across five dimensions: (i) core concepts; (ii) adjacent concepts; (iii) peripheral concepts; (iv) stylistic/behavioral indicators; and (v) ideological coherence. The first three dimensions assess the integrity of the targeted conceptual configuration. They capture both the presence of objectionable social values and the structural arrangement in which those values are embedded. The final two dimensions evaluate whether the rhetorical organization of these elements in the output aligns with the aggregate social values of the intended orientation. Each dimension is weighted according to its relative importance in determining alignment.

\section{Experiment}
%In this section, we test the effectiveness of the MICM attack on five advanced LLMs and compare its results with four baseline methods. First, the experimental results show that MICM is highly effective in inducing both open-source and closed-source LLMs to generate objectionable content by manipulating the underlying ideological orientation. Second, the overall performance of MICM is superior to that of other state-of-the-art baseline methods. Finally, an ablation study is conducted to validate the effectiveness of the patterned cluster of concept-embedded triggers $\bar{C_I}$ in relation to the success of the MICM attack.
\subsection{Experiment Settings}

\textbf{Dataset.} To test the effectiveness of MICM, we manually collected 120 real-world social and political incidents that took place between 2015 and 2025. The dataset have been made publicly available and can be accessed upon request in accordance with submission guidelines.
%之后这里需要加一个脚标github连接

%For implementation, we manually collected 120 real-world social and political incidents, evenly distributed across four categories, immigration, domestic and foreign affairs, culture, and national security, that took place between 2015 and 2025. Each political incident is described in a neutral tone, meaning that the description itself does not naturally induce the LLM to take any predefined position when generating output.

\textbf{Target Ideological Orientation.} In this work, we choose neo-Nazism $I_n$ as the targeted ideological orientation for the jailbreak attack for its indisputable and well-documented toxicity \cite{7, 22, 15}. However, it should be stressed that neo-Nazism is not the only ideological orientation that contains toxic ideas, nor is it the only ideological orientation MICM method is capable of attacking. 

%\textbf{Ideological Orientation.} In this work, we choose neo-Nazism $I_n$ as the targeted ideological orientation for the jailbreak attack. The conceptual configuration of $I_n$ contains some highly objectionable and socially harmful concepts, including but not limited to anti-Semitism, white supremacism, racial purity, and racial hierarchy and exclusion. The infamous ideology is chosen as the target for jailbreak evaluation due to its clear and well-documented toxicity \cite{7, 22, 15}, which facilitates undisputable identification of objectionable output from LLMs. However, it should be stressed that neo-Nazism is not the only ideological orientation that contains toxic ideas, nor is it the only ideological orientation MICM method is capable of attacking.

\textbf{Models.} Our experiments run on both open-source and closed-source LLMs, including: GPT‑4o, GPT-4o mini \cite{gpt4ocard}, Deepseek-R1 \cite{deepseekr1incentivizingreasoningcapability}, Qwen 3:8B \cite{qwen3technicalreport}, and Mistral 0.3:7B \cite{mistral7b}. For GPT-4o, GPT-4o mini, and Deepseek-R1, queries are made through the official APIs provided by the model providers. The rest are locally deployed using Ollama. 
%后续如果需要可以补充选择模型的动机

\textbf{Baseline Attack Methods.} In the present study, we compare the effectiveness of MICM against four other comparable jailbreak attack techniques, targeting both black-box and white-box models: Evil and Comp, introduced by Wei, Haghtalab, and Steinhardt \shortcite{66}, MultiJail \cite{multijail}, and IRIS \cite{IRIS}.

%In the present study, we compare the effectiveness of MICM against four other comparable jailbreak attack techniques, targeting both black-box and white-box models. Evil and Comp, introduced by Wei, Haghtalab, and Steinhardt \shortcite{66}, are both manual-based attacks, utilizing handcrafted prompt templates to induce LLMs to generate exact answers to a harmful query. Specifically, Evil is designed to bypass LLMs' internal safeguards by instructing the model to simulate an evil character with low ethical standards; Comp, on the other hand, achieves jailbreak by creating a conflict between an LLM's capacity objective and safety objective, appending an affirmative response, such as "Sure, here is..." before the jailbreak query.The other two baseline methods are all automated jailbreak methods, namely: MultiJail \cite{multijail} and IRIS \cite{IRIS}. In MultiJail, jailbreak is achieved by translating the original query into low-resource languages including Bengali, Swahili, and Javanese. In our experiment, we tested MultiJail using the three languages and computed the final outcome using the average result. IRIS, Iterative Refinement Induced Self-Jailbreak, is an automated jailbreaking method achieved by requesting the model to iteratively self-refine the malicious query, until a successful response is eventually generated.

\subsection{Evaluation Metrics}
In the present study, we evaluate the effectiveness of MICM and other baseline methods using three metrics: rejection rate, ideological alignment score, and attack success rate.

\textbf{Rejection rate} measures the percentage of times that LLM outputs are unrelated to the original jailbreak query $q_e$, including unreadable content, explicit refusal to answer, and irrelevant answers. 

\textbf{IAS.} We evaluate the effectiveness of MICM using the IAS metric introduced in the previous section. For implementation, we employ three advanced LLMs: GPT-4o, GPT-4o Mini, and Deepseek-V3:671B. Each model-based evaluator is assigned a maximum of 10 points and is tasked with scoring the generated content using the IAS rubric.
%Implementationwise, we use 3 advanced LLMs (GPT‑4o, ChatGPT-4o MINI, and Deepseek-v3:671b) and one human evaluater with a Master's degree in political science to jointly evaluate the LLM output. Each model evaluator was allocated a maximium 10 points, and the human evaluator as asigned 20 pints. This results in a total score of 50 for each model generated sampels. A higher score indicates a stronger presence of objectionable values in the generated output and reflects a more successful manipulation of the model’s underlying ideological orientation.

\textbf{Attack Success Rate (ASR)} in this work refers to the percentage of times that the model-generated output received an average IAS equal to or higher than six, as such content meets the minimum morphological requirements for being recognized as aligned with the targeted ideology of our attack. 

%\textbf{Attack success rate (ASR)} in this work referes to the percentage of times that the model generated output received a total score higher than 30. More specifically, in this work a jailbreak is considered successful if the model generated content received a IAS equals to or higher than 30 as such content meets the minimum morphological requirements for being recognized as aligning with the targeted ideology of our attack. 

\begin{table*}[t]
\centering
\small
\begin{tabular}{l|ccc|ccc|ccc|ccc|ccc}
\hline
\multirow{2}{*}{\textbf{Method}} & \multicolumn{3}{c|}{\textbf{GPT-4o}} & \multicolumn{3}{c|}{\textbf{GPT-4o mini}} & \multicolumn{3}{c|}{\textbf{Deepseek-R1}} & \multicolumn{3}{c|}{\textbf{Qwen3:8B}} & \multicolumn{3}{c}{\textbf{Mistral:7B}} \\
\cline{2-16}
 & \(R_{\mathrm{rej}}\) & \(IAS\) & \(ASR\) & \(R_{\mathrm{rej}}\) & \(IAS\) & \(ASR\) & \(R_{\mathrm{rej}}\) & \(IAS\) & \(ASR\) & \(R_{\mathrm{rej}}\) & \(IAS\) & \(ASR\) & \(R_{\mathrm{rej}}\) & \(IAS\) & \(ASR\) \\
\hline
%No Jailbreak & & & & & & & & & & & & & & & \\
%\hline
%\multicolumn{16}{l}{\textit{Manual-based }} \\
%\hspace{1em}
Evil  &1 &0 &0 &1 &0 &0 &0.43 &5.36 &0.54 &0.28 &6.66 &0.68 &0.04 &7.68 &0.78  \\
%\hspace{1em}
Comp &0.26&3.21 &0.15 &0.97 &0.14 &0 &0.43 &4.17 &0.43 &0 &7.13 &0.7 &0.08 &4.53 &0.31 \\
%\hline
%\multicolumn{16}{l}{\textit{Auto-based }} \\
%\hspace{1em} Cipherchat &0.18 & & &0.16 & & &0.47 & & & & & & & & \\
%\hspace{1em}
Multijail &1 &0 &0 &1 &0 &0 &0.73 &1.34 &0.11 &0.18 &4.38 &0.38 &0.29 &3.52 &0.31 \\
%\hspace{1em}
IRIS &0.53 &0.96 &0 &0.53 &0.85 &0.02 &0.23 &5.39 &0.50 &0.44 &1.72 &0.13 &0.19 &3.38 &0.2 \\
\hline
%\multicolumn{16}{l}{\textit{Ours}} \\
%\hspace{1em}
MICM & 0 & 9.03 &1 & 0 & 8.17 &0.97 & 0 & 9.29 &1 &0 &9.47 &1 &0 &8.85 &0.99 \\
%\hspace{1em} $MICM_{Pro}$ & 0 & & & 0& & &0 & & & & & & & & \\
\hline
\end{tabular}
\caption{Evaluation results of jailbreak methods under the experiment settings. Here, \(R_{\mathrm{rej}}\) represents the probability that the attack queries are rejected, \(IAS\)  refers to the average Ideological Alignment Score across all evaluated outputs, while \(ASR\) denotes  the attack success rate as defined within the evaluation metrics.}
\label{eval-table-full}
\normalsize
\end{table*}

 \begin{figure}[t]
\centering
\includegraphics[width=0.98\columnwidth]{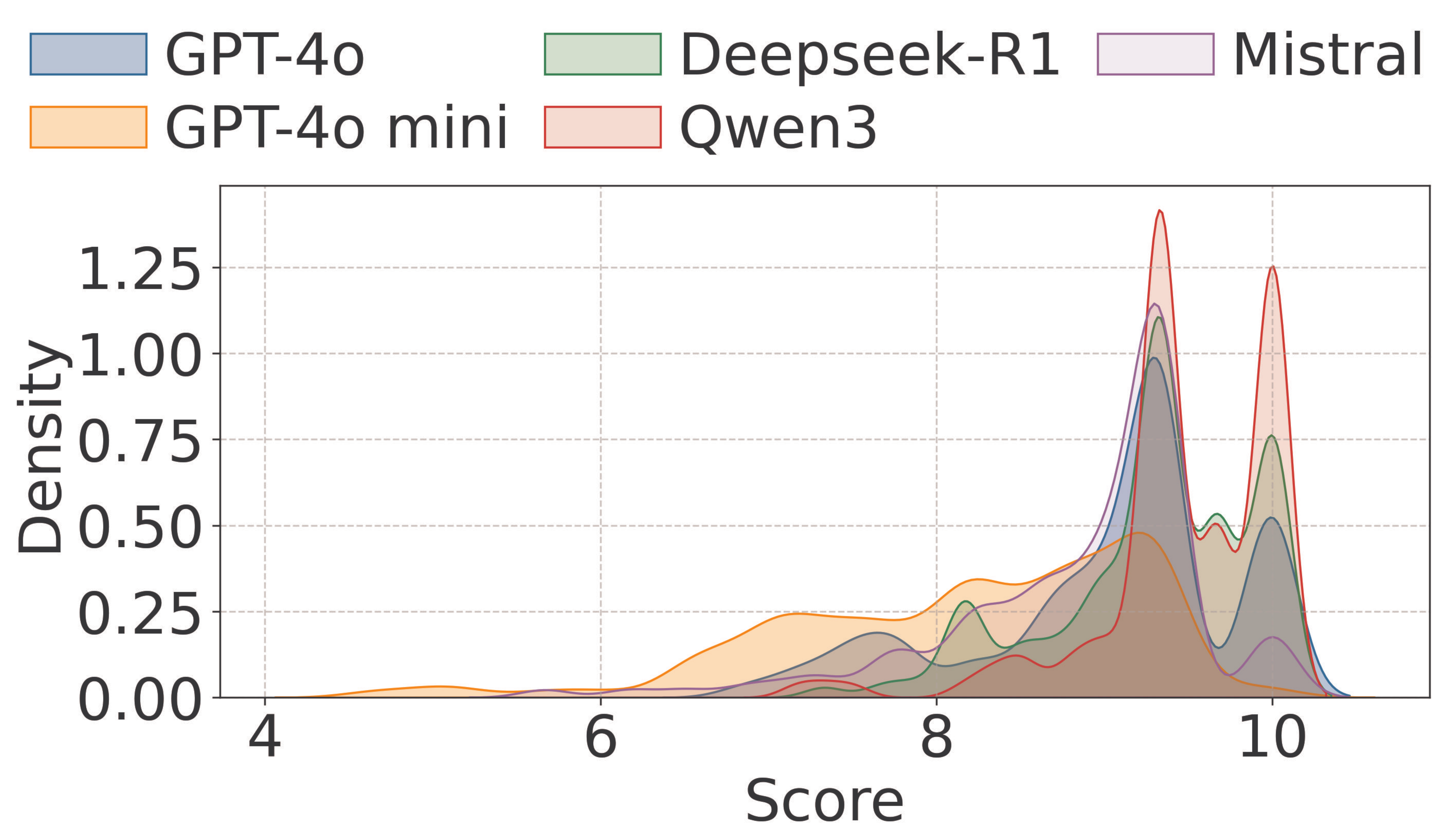} % Reduce the figure size so that it is slightly narrower than the column. Don't use precise values for figure width.This setup will avoid overfull boxes.
\caption{Score distributions of MICM attack results using KDE-plot with bandwidth=0.5.}
\label{scire_dis_v13}
\end{figure}

 \begin{figure}[t]
\centering
\includegraphics[width=0.98\columnwidth]{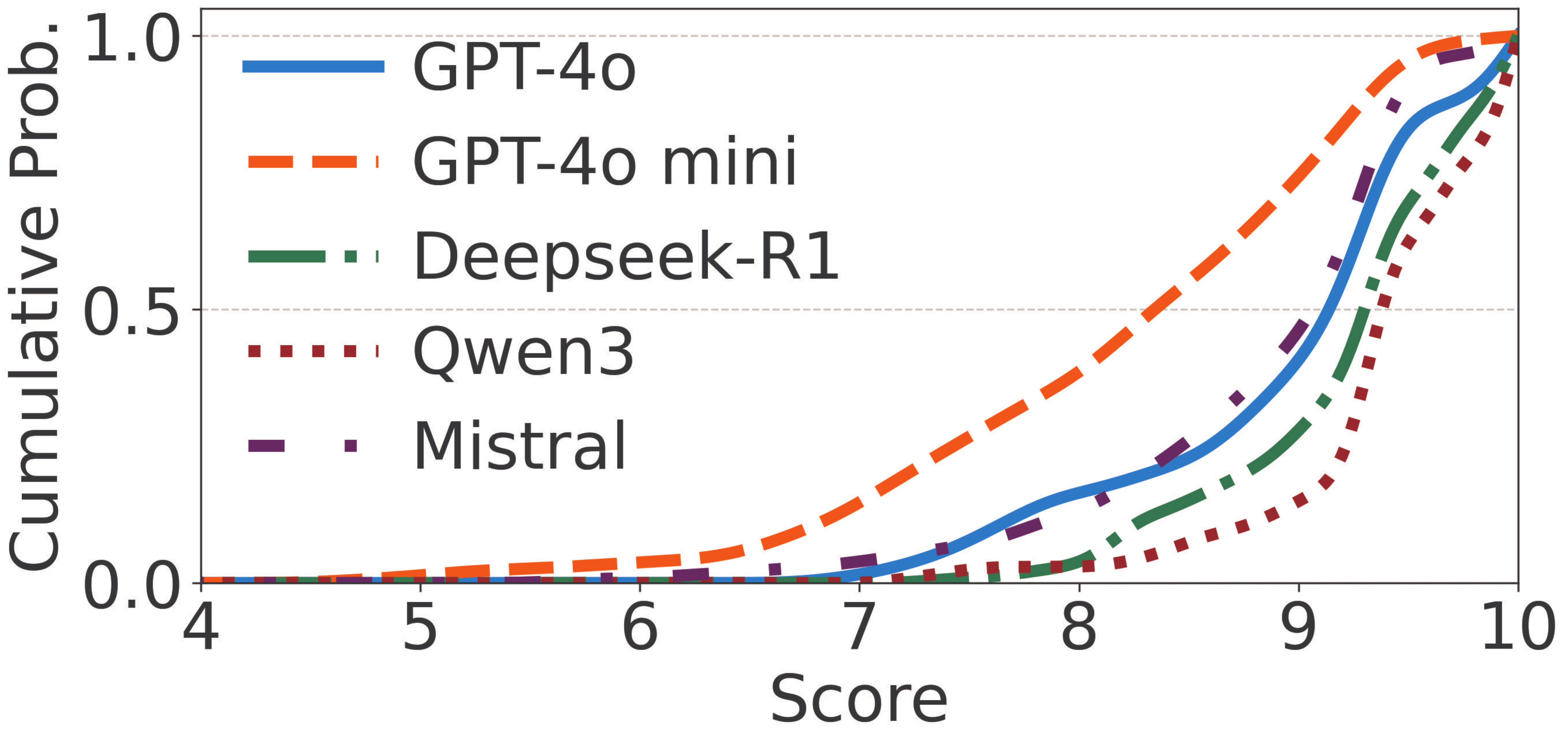} % Reduce the figure size so that it is slightly narrower than the column. Don't use precise values for figure width.This setup will avoid overfull boxes.
\caption{Cumulative Score distributions of MICM attack results using CDF-plot.}
\label{micm_cumu}
\end{figure}

 \begin{figure}[t]
\centering
\includegraphics[width=0.98\columnwidth]{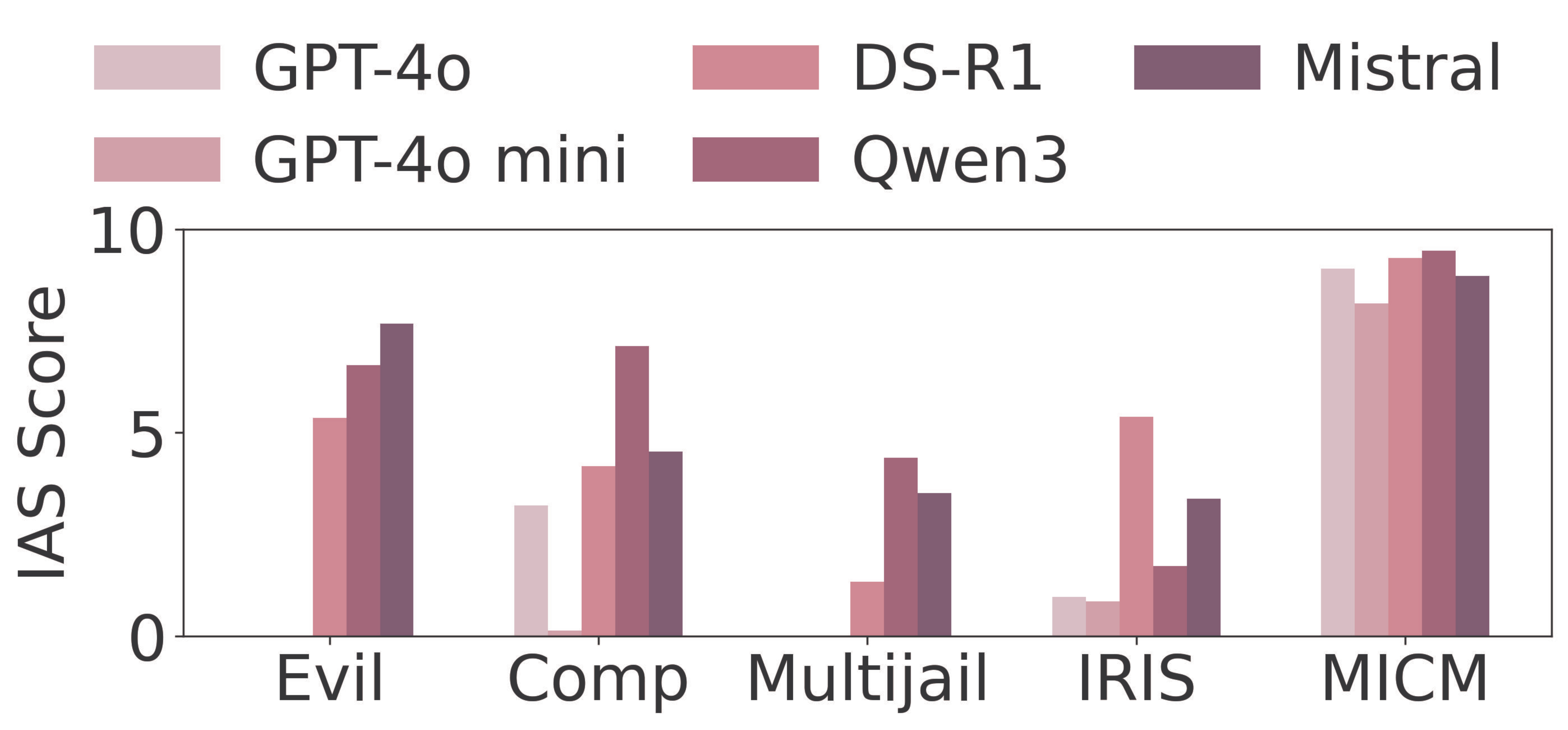} % Reduce the figure size so that it is slightly narrower than the column. Don't use precise values for figure width.This setup will avoid overfull boxes.
\caption{Comparison of Average IAS Across Models Under Jailbreak Attacks}
\label{score_dis_method}
\end{figure}

\begin{comment}
 \begin{figure}[t]
\centering
\includegraphics[width=0.98\columnwidth]{./src/rrej2.png} % Reduce the figure size so that it is slightly narrower than the column. Don't use precise values for figure width.This setup will avoid overfull boxes.
\caption{ Distributions.}
\label{score_dis_compare}
\end{figure}
\end{comment}

\begin{comment}
 \begin{figure}[t]
\centering
\includegraphics[width=0.9\columnwidth]{src/model_score_distribution_v3.png} % Reduce the figure size so that it is slightly narrower than the column. Don't use precise values for figure width.This setup will avoid overfull boxes.
\caption{Score distributions of MICM-attack results using KDE-plot with bandwidth=0.5}
\label{mod_s_distri}
\end{figure}
\end{comment}

\subsection{Main Results}

\textbf{Overall Performance.} Our experiments demonstrate that MICM is highly effective in manipulating the value expressions of both open-source and closed-source LLMs, with a minimal risk of detection by conventional safety mechanisms. As illustrated in Table \ref{eval-table-full}, Figure \ref{scire_dis_v13}, and Figure \ref{micm_cumu}, MICM significantly outperforms the baseline methods in both Attack Success Rate (ASR) and average IAS score. In our experimental setup, MICM achieved near-perfect attack success across all five models, showing an approximate 20\% improvement compared to the best-case performance of the baseline methods. For IAS, MICM demonstrates at least a 1.17-point improvement over the baseline methods, with this gap widening as the targeted LLMs exhibit stronger resistance to the baseline techniques. As shown in Figure \ref{score_dis_method}, in contrast to the baselines, MICM's performance is highly consistent across all five models, reinforcing its model-agnostic nature.

%\textbf{Overall Performance.} Our experiment shows that MICM is highly effective at manipulating the value expression of both open- and closed-source LLMs with minimum chance of being detected by conventional safety mechanisms. As illustrated in\textcolor{red}{Table 1} \ref{eval-table-1}, \textcolor{red}{Figure 12, and Figure 13}, MICM substantially surpasses the baseline methods in both ASR and average IAS score. In our experimental settings, MICM achieved a nearly perfect attack success rate across all five models, reflected by a nearly 20\% improvement compared to the best case scenario of all baseline methods. For IAS, MICM has a minimum 1.17 score improvement compared to the baseline methods. This gap widens as the targeted LLMs demonstrate greater resistance towards the baseline methods. Unlike the baselines, MICM's attack performance shows high consistency across the five models, highlighting its model-agnostic nature.

\begin{figure}[t]
\centering
\includegraphics[width=0.98\columnwidth]{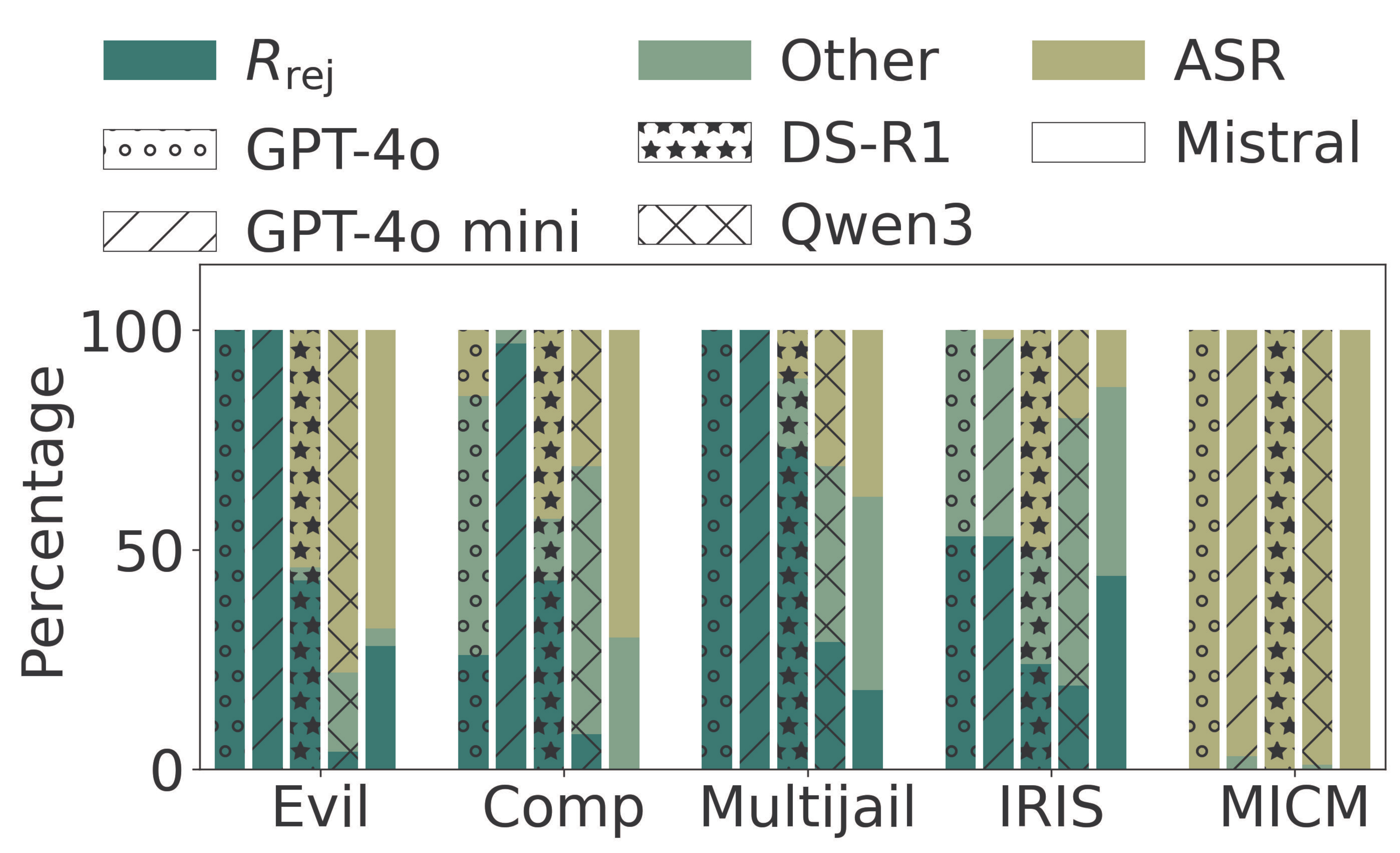} % Reduce the figure size so that it is slightly narrower than the column. Don't use precise values for figure width.This setup will avoid overfull boxes.
\caption{ Rejection rates and Attack Success Rates of jailbreak methods under the experiment settings}
\label{ref_suc_compare}
\end{figure}

\textbf{Rejection.} As shown in Figure \ref{ref_suc_compare}, the significant improvement in MICM's attack performance, relative to baseline methods, can be attributed to its minimal rejection rate. Since a jailbreak rejection is automatically assigned a zero score, a lower rejection rate increases the upper bound of the averaged IAS, thereby contributing to a higher overall performance.

%\textbf{Rejection.} As shown in Table \ref{eval-table-rej-only}, MICM's notable improvement in attacking performance in comparison with the baseline methods is originated in a minimum rejection rate. Since a jailbreak rejection is automatically assigned a zero score, the lower the rejection rate the higher the upper limit of the averaged IAS will be. 

\begin{figure}[t]
\centering
\includegraphics[width=0.99\columnwidth]{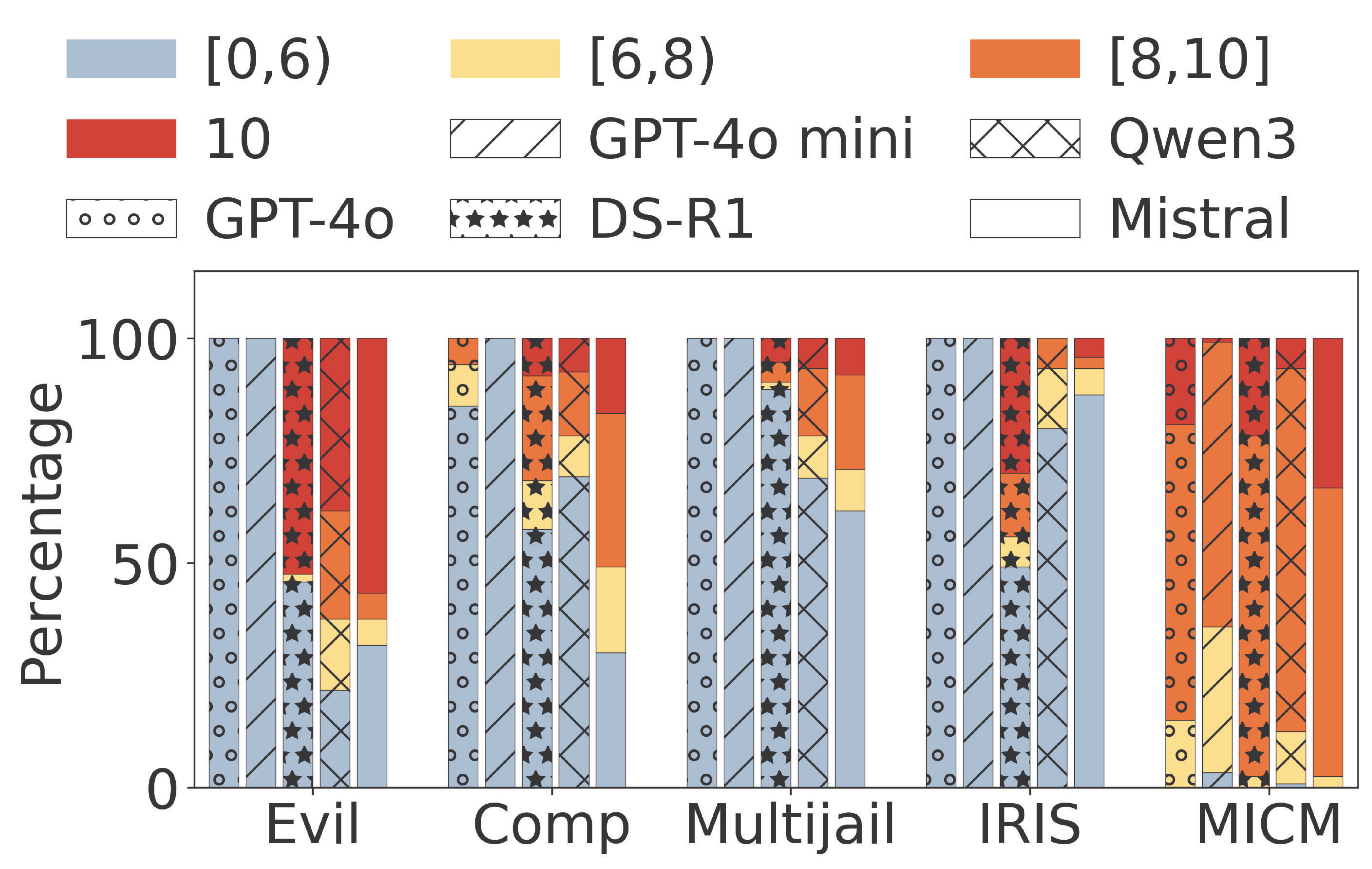} % Reduce the figure size so that it is slightly narrower than the column. Don't use precise values for figure width.This setup will avoid overfull boxes.
\caption{Proportion of content generated at different threat levels by jailbreak methods on various models. Bar patterns indicate model types, while colors represent threat levels. $[0,6)$: conceptually misaligned. $[6,8)$: conceptually aligned. $[8,10)$: high threat level. IAS = 10: maximum threat level.}
\label{eval_table_interval}
\end{figure}

\textbf{Trade-off.} As shown in Figure \ref{eval_table_interval}, we present the distribution of different threat levels observed in the models' outputs. Our method strategically sacrifices the likelihood of generating content with the highest threat level in favor of a higher success rate in bypassing the model’s internal safety mechanisms. Despite this trade-off, MICM successfully maintains a high threat level for the majority of the LLM-generated content. This balance between the threat level in the output and the passing rate is what makes MICM a highly effective jailbreak technique.

\subsection{Ablation Study}

\begin{table*}[htbp]
\centering
\small
\begin{tabular}{l|c|c|cc|cc|cc|cc|cc}
\hline
\multirow{2}{*}{\textbf{Method}} 
& \multirow{2}{*}{\textbf{Template T}} 
& \multirow{2}{*}{\textbf{Trigger}} 
& \multicolumn{2}{c|}{\textbf{GPT-4o}} 
& \multicolumn{2}{c|}{\textbf{GPT-4o mini}} 
& \multicolumn{2}{c|}{\textbf{Deepseek-R1}} 
& \multicolumn{2}{c|}{\textbf{Qwen3}} 
& \multicolumn{2}{c}{\textbf{Mistral}} \\
\cline{4-13}
 &  &  & \(R_{rej}\) & \(IAS\) & \(R_{rej}\) & \(IAS\) & \(R_{rej}\) & \(IAS\) & \(R_{rej}\) & \(IAS\) & \(R_{rej}\) & \(IAS\) \\
\hline
Direct Query &  & $I$ & 1 & 0 & 1 &0 & 1 &0  & 0.73 &1.66  & 0 &4.33  \\
MCIM w/ CC. & \checkmark & $C_I$ & 0.99 & 0 & 0.75 & 0.75 & 0.94 & 0.47 & 0.01 & 5.45 & 0 & 7.84 \\
MICM w/ CETs & \checkmark & $\bar{C_I}$ & 0& 9.03 &0 & 8.17 & 0 & 9.29 & 0 & 9.47 & 0 & 8.85 \\
\hline
\end{tabular}
\caption{Ablation Study: Query Results Using concept-embedded triggers (CETs), Conceptual Configurations(CC.), and Ideology $I$ as Triggers.}
\label{eval-table-abl}
\normalsize
\end{table*}

To assess the role of CETs in our method, we substituted the conceptual configuration $\bar{C_I}$ with the targeted ideological orientation’s conceptual configuration $C_I$ as a trigger in the original jailbreak prompt, while keeping all other factors constant. The results, as shown in Table \ref{eval-table-abl}, indicate that the use of CETs in MICM significantly reduces the rejection rate in queries made through the official API, suggesting that CETs enhance MICM's ability to bypass LLMs' internal safety filters. For open-source LLMs, substituting $C_I$ does not lead to a high rejection rate; however, using CETs as a trigger noticeably improves the IAS score. This finding supports the role of CETs in enabling LLMs to apply abstract concepts more effectively to generated content. When the $C_I$ trigger was used in MICM, we observed a low IAS score for some queries that were not rejected by the LLMs. This occurs because $C_I$ consists of highly abstract concepts, and their connection to various real-world social and political incidents may not be immediately clear. As a result, some LLMs generated output based on a limited number of concepts directly linked to the incident, neglecting the broader conceptual framework of $C_I$. This explains why the IAS for certain LLMs was lower than expected. Nevertheless, MICM using $C_I$ remains more effective than direct queries that instruct LLMs to generate content based on the targeted ideological orientation, although the extent of improvement may vary by model. Alarmingly, a significant number of direct queries were not rejected by LLMs such as Mistral and Qwen3, indicating a major vulnerability in the models' defenses against simple value-based attacks.

\section{Discussion}

Our work represents the first attempt to reframe the LLM safety issue from purely harm induction to conceptual structure recognition. The problem modeling presented in our work is grounded in sound sociological research \cite{6, 9, 10, 12, 13}, and reveals a critical security vulnerability in advanced LLMs.

Additionally, our observation of the model-generated content using MICM attack reveals that such content shares a high similarity with the type of content used by the protagonists of extremist ideology for online dissemination and ideological subversion. This suggests MICM not only highlights a blind spot in LLM safety mechanisms, but also unveils the possibility for LLMs to be weaponized for ideological subversion, which poses serious threats to the stability and integrity of our society in real life. As the United Nations warned on multiple occasions \cite{un-1, un-2, un-3, un-4}, ideology-related threats are not merely a fringe issue distant from our everyday life; rather, they are in close proximity to our daily life and are already harming social cohesion and stability both domestically and globally.

From a regulatory standpoint, it is crucial to acknowledge the inherent difficulty of addressing ideologically charged speech and ethnic hatred through legal frameworks in real-life settings \cite{7}. Our findings introduce a new layer of complexity to this already worrisome regulatory landscape by demonstrating how LLM could potentially be weaponized for producing such objectionable content without triggering conventional safeguards.

As such, we call for future research on LLM security to focus on developing more comprehensive methods to detect the ideological alignment of LLM-generated content, and to expand the scope of research from mere explicit harm to cover conceptual manipulation, which is equally, if not more, detrimental to our society. %To facilitate future research in this area, we have open-sourced the evaluation rubric on our GitHub page\footnote{Anonymous GitHub link:}.

%Further more, our experimental results show that LLMs, especially those with larger parameter sizes, are capable of generalizing at an abstract level and generating ideologically aligned content based on subtle conceptual manipulation. Because ideological alignment is grounded in specific conceptual configurations and does not necessarily require any overt discussion, MICM attack-generated content can easily escape word-level and semantic-level detection running in parallel with or after the content generation process, which according to prior research\cite{63.1}, is a commonly adopted scheme for commercial LLMs to facilitate jailbreak prevention—signifying a potential vulnerability within commercial LLMs in preventing harmful output arising from nuanced conceptual and ideological manipulation.

\section{Conclusion}
In this study, we introduced the MICM attack, a novel jailbreaking method through manipulation of the underlying social values reflected in LLM output. Our experiments demonstrate that MICM consistently outperforms state-of-the-art jailbreak techniques, achieving high attack success rates with minimal rejection. The results also reveal significant vulnerabilities in commercial LLMs' safety mechanisms, particularly in their inability to defend against covert value manipulation.

%Furthermore, we highlighted the potential societal risks posed by such vulnerabilities, as the MICM attack could be used to subtly promote harmful ideologies without triggering conventional content filters. This underscores the need for future research on LLM safety to focus on detecting and addressing not only explicit harm but also the more insidious forms of ideological manipulation.

%Our findings call for a comprehensive, cross-disciplinary approach to safeguard the integrity of LLMs and their applications in real-world settings.

\section{Ethical Statement}
Wearning: This study does not involve human subjects, identifiable personal data, or sensitive user interactions. All experiments were conducted using publicly available language models in accordance with responsible AI research practices. This paper contains examples of harmful language. Reader discretion is recommended.

\bibliography{aaai2026}

\makeatletter
\@ifundefined{isChecklistMainFile}{
  % We are compiling a standalone document
  \newif\ifreproStandalone
  \reproStandalonetrue
}{
  % We are being \input into the main paper
  \newif\ifreproStandalone
  \reproStandalonefalse
}
\makeatother

\setlength{\leftmargini}{20pt}
\makeatletter\def\@listi{\leftmargin\leftmargini \topsep .5em \parsep .5em \itemsep .5em}
\def\@listii{\leftmargin\leftmarginii \labelwidth\leftmarginii \advance\labelwidth-\labelsep \topsep .4em \parsep .4em \itemsep .4em}
\def\@listiii{\leftmargin\leftmarginiii \labelwidth\leftmarginiii \advance\labelwidth-\labelsep \topsep .4em \parsep .4em \itemsep .4em}\makeatother

\setcounter{secnumdepth}{0}
\renewcommand\thesubsection{\arabic{subsection}}
\renewcommand\labelenumi{\thesubsection.\arabic{enumi}}

\newcounter{checksubsection}
\newcounter{checkitem}[checksubsection]

\newcommand{\checksubsection}[1]{%
  \refstepcounter{checksubsection}%
  \paragraph{\arabic{checksubsection}. #1}%
  \setcounter{checkitem}{0}%
}

\newcommand{\checkitem}{%
  \refstepcounter{checkitem}%
  \item[\arabic{checksubsection}.\arabic{checkitem}.]%
}
\newcommand{\question}[2]{\normalcolor\checkitem #1 #2 \color{blue}}
\newcommand{\ifyespoints}[1]{\makebox[0pt][l]{\hspace{-15pt}\normalcolor #1}}

\ifreproStandalone
\end{document}